\documentclass{article}
\usepackage{spconf,amsmath,graphicx}
\usepackage{algorithm}
\usepackage[]{graphicx}
\usepackage{stfloats}
\usepackage{cite}
\usepackage{psfrag}
\usepackage{wrapfig}
\usepackage{epsfig}
\usepackage{url}
\usepackage{amsmath, amssymb, bm}
\usepackage{array}
\usepackage{times}
\usepackage{multicol}
\usepackage{algorithm}
\usepackage{algorithmic}
\usepackage{mathrsfs}
\usepackage{multirow}
\usepackage{txfonts}
\usepackage[normalem]{ulem}

\usepackage[utf8x]{inputenc}
\usepackage{ucs}
\usepackage{makeidx}
\usepackage[dvipsnames,usenames]{color}
\usepackage{array}
\usepackage{colortbl}
\usepackage{booktabs}
\usepackage{color}
\usepackage{cleveref}
\usepackage{setspace}
\usepackage{subcaption}
\usepackage{enumitem} 
\usepackage{soul} 
\usepackage{todonotes} 
\usepackage{balance}

\renewcommand{\vec}[1]{\mathbf{#1}}
\newcommand{\setsym}[1]{\mathbb{#1}}

\graphicspath{ {./fig/} }

\title{Advancing Acoustic-to-Word CTC Model}
\name{Jinyu Li, Guoli Ye, Amit Das\sthanks{Work performed during an internship at Microsoft.}, Rui Zhao, Yifan Gong}
\address{Microsoft AI and Research, One Microsoft Way, Redmond, WA 98052\\
}
\begin{document}
\ninept
\maketitle
\begin{abstract}
The acoustic-to-word model based on the connectionist temporal classification (CTC) criterion was shown as a natural end-to-end (E2E) model directly targeting words as output units. However, the word-based CTC model suffers from the out-of-vocabulary (OOV) issue as it can only model limited number of words in the output layer and maps all the remaining words into an OOV output node. Hence, such a word-based CTC model can only recognize the frequent words modeled by the network output nodes. Our first attempt to improve the acoustic-to-word model is a hybrid CTC model which consults a letter-based CTC when the word-based  CTC model emits OOV tokens during testing time. Then, we propose a much better solution by training a mixed-unit CTC model which decomposes all the OOV words into sequences of frequent words and multi-letter units. Evaluated on a 3400 hours Microsoft Cortana voice assistant task, the final acoustic-to-word solution improves the baseline word-based CTC by relative 12.09\% word error rate (WER) reduction when combined with our proposed attention CTC. Such an E2E model without using any language model (LM) or complex decoder outperforms the traditional context-dependent phoneme CTC which has strong LM and decoder by relative 6.79\%.
\end{abstract}
\vspace{-2mm}
\section{Introduction}
\label{sec: Introduction}\vspace{-2mm}

As one of the most popular end-to-end (E2E) methods, the connectionist temporal classification (CTC) approach \cite{Graves-CTCFirst, Graves-E2EASR} was introduced to map the speech input frames into an output label sequence \cite{Hannun-DeepSpeech, sak2015learning, sak2015fast, miao2015eesen, kanda2016maximum, soltau2016neural, Zweig-AdvancesNeuralASR, liu2017gram, audhkhasi2017direct, Li17CTCnoOOV, Yu-RecentProgDeepLearningAcousticModels, Li2018Speaker}. To deal with the issue that the number of output labels is smaller than that of input speech frames in speech recognition tasks, CTC introduces a special blank label and allows for repetition of labels to force the output and input sequences to have the same length. 

CTC outputs are usually dominated by blank symbols and the output tokens corresponding to the non-blank symbols usually occur with spikes in their posteriors. Thus, an easy way to generate ASR outputs using CTC is to concatenate the non-blank tokens corresponding to the posterior spikes and collapse those tokens into word outputs if needed. This is a very attractive feature for E2E  modeling as there is neither LM nor complex decoding involved. We refer this decoding strategy as greedy decoding, and our E2E models studied in this paper all use greedy decoding. 

As the goal of ASR is to generate a word sequence from speech acoustics, word is the most natural output unit for network modeling. 
A big challenge in the word-based CTC is the out-of-vocabulary (OOV) issue \cite{bazzi2002modelling, decadt2002transcription, yazgan2004hybrid, bisani2005open}. In \cite{sak2015fast, soltau2016neural, audhkhasi2017direct}, only the most frequent words in the training set were used as targets whereas the remaining words were just tagged as OOVs. All these OOV words can neither be further modeled nor be recognized during evaluation.  For example in \cite{sak2015fast}, the CTC with up to 27 thousand (k) word output targets was explored but the ASR accuracy is not very good, partially due to the high OOV rate when using only around 3k hours training data.  

To solve this OOV issue in the word-based CTC, we proposed a hybrid CTC  \cite{Li17CTCnoOOV} which uses the output from the word-based CTC as the primary ASR result and consults a letter-based CTC at the segment level where the word-based CTC emits an OOV token. A shared-hidden-layer structure is used to align the word segments between the word-based CTC and the letter-based CTC so that the OOV token lookup algorithm can work.  
However, the shared-hidden-layer structure still cannot guarantee a perfect alignment between the word and letter based CTCs. It also hurts the modeling accuracy of the auxiliary CTC model. In \cite{audhkhasi2017building}, a spell and recognize model is used to learn to first spell a word and then recognize it. Whenever an OOV is detected, the decoder consults the letter sequence from the speller. In \cite{Li17CTCnoOOV, audhkhasi2017building}, the displayed hypothesis is more meaningful than OOV to users. However, both methods cannot improve the overall recognition accuracy too much. 

In this study, we propose a solution to the OOV issue in the acoustic-to-word modeling by decomposing the OOV word into a  mixed-unit sequence of frequent words and letters at the training stage. We use attention CTC to address the inherent CTC modeling issue. During testing, we do greedy decoding for the whole E2E system in a single step without the need of using the two-stage (OOV-detection and then letter-sequence-consulting) process as in  \cite{Li17CTCnoOOV, audhkhasi2017building}.   With all these components, the final acoustic-to-word solution improves the baseline acoustic-to-word CTC by relative 12.09\% word error rate (WER) reduction and also outperforms the traditional context-dependent-phoneme CTC with strong LM and decoder by relative 6.79\%.

\vspace{-4mm}
\section{Advance Acoustic-to-Word CTC}\vspace{-2mm}
\label{sec: E2E}


\subsection{Word-based Connectionist Temporal Classification (CTC)}\vspace{-2mm}
\label{ssec: CTC}
A CTC network uses an recurrent neural network (RNN) and the CTC error criterion \cite{Graves-CTCFirst, Graves-E2EASR}  which directly optimizes the prediction of a transcription sequence. As the length of output labels is shorter than the length of input speech frames, a CTC path is introduced to have the same length as the input speech frames by adding the blank symbol as an additional label and allowing repetition of labels. 

Denote $\bf{x}$ as the speech input sequence, $\bm\pi$ as the CTC path, $\bf{l}$ as the original label sequence (transcription), and  $B^{-1}(\bf{l})$ as the preimage mapping all possible CTC paths $\bm\pi$ resulting from $\bf{l}$. Then, the CTC loss function is defined as the sum of negative log probabilities of correct labels as,
\vspace{-2mm}
\begin{equation}
L_{CTC} = - \ln P( {\bf{l}|\bf{x}} ) = - \ln \sum_{{\bm\pi} \in B^{-1}(\bf{l})} P( {\bm\pi}  | \bf{x} ).
\end{equation}
With the conditional independence assumption, $P( {\bm\pi} | \bf{x} )$ can be decomposed into a product of posteriors from each frame as,
\vspace{-2mm}
\begin{equation}
P( {{\bm\pi}  | \bf{x}} ) = \prod_{t=1}^T P( \pi_{t}| \bf{x}).
\end{equation}

As the goal of ASR is to generate a word sequence from the speech waveform, the word unit is the most natural output unit for network modeling. The recently proposed acoustic-to-word models \cite{soltau2016neural, audhkhasi2017direct}, a.k.a. word-based CTC models, build multiple layer long short-term memory (LSTM) \cite{Hochreiter1997long, Graves2013speech, Sak2014long}  networks and use words as the network output units, optimized with the CTC training criterion. It is very simple to generate the word sequence with this word-based CTC model using greedy decoding: pick the words corresponding to posterior spikes to form the output word sequence. There is neither language model nor complex decoding process involved.

However, when training a word-based CTC model, only the most frequent words in the training set were used as targets whereas the remaining words were just tagged as OOVs. All these OOV words cannot be modeled by the network and cannot be recognized during evaluation. For example, if the transcription of an utterance is ``have you been to newyorkabc'' in which newyorkabc is an infrequent word, the training token or recognition output sequence for this utterance will be ``have you been to OOV''.

\vspace{-2mm}
\subsection{Hybrid CTC}\vspace{-2mm}
\label{ssec: hybCTC}

To solve the OOV issue in the acoustic-to-word model, the hybrid CTC model uses a word-based CTC as the primary model and a letter-based CTC as the auxiliary model. The word-based CTC model emits a word sequence, and the output of the letter-based CTC is only consulted at the segment where the word-based CTC emits an OOV token. The detailed steps for building the hybrid CTC model are described as follows:
\begin{itemize}
	\item Build a multi-layer LSTM-CTC model with words as its output units. Map all the words occurring less than $N$ times in the training data as the OOV token. The output units in this LSTM-CTC model are all the words occurring at least $N$ times in the training data, together with OOV, blank, and silence tokens.
	\item Freeze the bottom $L-1$ hidden layers of the word-CTC, add one LSTM hidden layer and one softmax layer to build a new LSTM-CTC model with letters as its output units. 
	\item During testing, generate the word output sequence using greedy decoding. If the output word sequence contains an OOV token,  replace the OOV token with the word generated from the letter-based CTC that has the largest time overlap with the OOV token.
\end{itemize}

\vspace{-2mm}
\subsection{CTC with Multi-letter Units}\vspace{-2mm}
\label{ssec: multiCTC}

In \cite{Li17CTCnoOOV}, the letter-based CTC uses single-letter units as the output units. Inspired by gram CTC \cite{liu2017gram} and multi-phone CTC \cite{siohan2017ctc}, we extend the output units with double-letter and triple-letter units to benefit from long temporal units which are more stable. We hope to improve the hybrid CTC system as the OOV token may be replaced by more precise words generated by the CTC with multi-letter units.

Gram CTC and multi-phone CTC are based on letter and phoneme respectively, but allow to output variable number of letters (i.e., gram) and phonemes at each time step. The units in gram CTC and multi-phone CTC are learned automatically with the modified forward-backward algorithm to take care of all the decompositions. Both of them need much more complicated decoding than greedy decoding when generating outputs. In contrast, we just simply decompose every word into a sequence of one or more letter units, with examples shown in the first three rows of Table \ref{Tab:units}. This decomposition is much simpler, without changing the CTC forward-backward process and can use the same greedy decoding as the CTC with single-letter units.

\begin{table}[!t]
\centering
\caption{Examples of how words are represented with different output units. ``Newyork" is a frequent word while ``newyorkabc" is an infrequent word. The word-based CTC treats ``newyork" as a unique output node and ``newyorkabc" as the OOV output node.}
\vspace{-3mm}
\begin{tabular}{l|c c}
   \hline

		Decomposition Type								&newyork      &newyorkabc     \\\hline
All words: single-letter						    &n e w y o r k      &n e w y o r k a b c  \\
All words: double-letter            &ne wy or k      &ne wy or ka bc   \\
All words: triple-letter            &new yor k      &new yor kab c   \\ 
All words: word											&newyork      &OOV	\\
OOVs only: single-letter						&newyork      &n e w y o r k a b c \\
OOVs only: word+single-letter 	&newyork 	&newyork a b c   \\
OOVs only: word+triple-letter 	&newyork 	&newyork abc   \\ \hline
\end{tabular}
\vspace{-2mm}
\label{Tab:units}
\end{table}

\vspace{-2mm}
\subsection{Acoustic-to-Word CTC with Mixed Units}\vspace{-2mm}
\label{ssec: mixCTC}

In hybrid CTC, the shared-hidden-layer constraint is used to help the time synchronization of word outputs between the word-based and letter-based CTC models. However, the blank symbol dominates most of the frames, and therefore the time synchronization is not very reliable. 
The ideal case should be when the spoken word is in the frequent word list  the system emits a word output. And when the spoken word is an OOV (infrequent) word, the system emits a letter sequence from which a word is generated by collapsing all those letters. This cannot be done with the hybrid CTC because the two CTCs are running in parallel without a perfect time synchronization. A direct solution is to train a single CTC model with mixed units. If the word is a frequent word, then we just keep it in the output token list. If the word is an infrequent word, then we decompose it into a letter sequence. As shown in the fifth row of Table  \ref{Tab:units}, the infrequent word ``newyorkabc" is decomposed into ``n e w y o r k a b c'' for single letter decompositions. However, the frequent word ``newyork" is not decomposed because it is a frequent word. Therefore, the output units of the CTC are mixed units, with both words (for frequent words) and letters (for OOV words). 

However we note that artificially  decomposing OOVs only into single-letter sequences may confuse CTC training because the network output modeling units are frequent words and letters.  To solve such a  potential issue, we decompose the OOV words into a combination of frequent words and letters. For example, in the last two rows of Table  \ref{Tab:units}, ``newyorkabc" is decomposed into ``newyork a b c'' if we use single-letter units with words or ``newyork abc'' if we use triple-letter units with words.  
In the CTC with mixed units, we use ``\$'' to separate each word in the sentence.  For example, the sentence ``have you been to newyorkabc'' is decomposed into ``\$ have \$ you \$ been \$ to \$ newyork abc \$''. If \$ is not used to separate words, we don't know how to collapse the mixed units (words+letters) into output word sequences.
Now, because during training the OOV words are decomposed into mixed units from words and letters, there is no OOV output node in the mixed unit CTC model. Consequently, during testing the model is very likely to emit OOV words as a sequence of frequent words and letters while still emitting frequent words when frequent words are spoken. 

\vspace{-2mm}
\subsection{CTC with Attention}\vspace{-2mm}
\label{ssec: CTCAttn}
We present a brief outline of modeling attention directly within CTC proposed by us in \cite{Das18CTCAttention}. One drawback of standard CTC training is the hard alignment problem. This is because CTC relies only on one hidden feature to make the current prediction. CTC Attention overcomes the hard alignment problem by producing a context vector which is a weighted sum of the most relevant hidden features within a context window. The resulting context vector can then be used to make the current prediction. Thus, the main components of proposed CTC Attention are: (a) the generation of context vectors as time convolution (TC) features, and (b) the computation of the weights of the hidden features using an attention mechanism. In this section, we use indices $t$ and $u$ to denote the time step for input and output sequences respectively. However, it is understood that in CTC every input frame $\vec{x}_{t}$ generates output $\vec{y}_{t} = \vec{y}_{u}$.

The context vector $\vec{c}_{u}$ can be computed as a TC feature by convolving the hidden feature $\vec{h}_{t}$ with learnable weight matrices $\vec{W}^{\prime}$ across time as,
\vspace{-2mm}
\begin{align}
\vec{c}_{u} &= \vec{W}^{\prime} \ast \vec{h} = \sum_{t =u-\tau}^{u+\tau} \vec{W}^{\prime}_{u - t} \vec{h}_{t} \nonumber \\
			&\stackrel{\Delta}{=} \sum_{t =u-\tau}^{u+\tau} \vec{g}_{t} = \gamma \sum_{t =u-\tau}^{u+\tau} \alpha_{u,t} \vec{g}_{t}. \label{eq:CTCAttn-TimeConvolution}
\end{align}
The duration $[u-\tau,\ u+\tau]$ represents a context window of length $C = 2\tau + 1$ and $\vec{g}_{t}$ represents the $filtered$ signal at time $t$. The last step in Eq. \eqref{eq:CTCAttn-TimeConvolution} holds when $\alpha_{u,t} = \frac{1}{C}$ and $\gamma = C$. The term $\alpha_{u,t}$ is the attention weight determining the relevance of $\vec{h}_{t}$ in generating $\vec{c}_u$. The context vector $\vec{c}_{u}$ is related to the output $\vec{y}_u$ using the softmax operation as,
\vspace{-2mm}
\begin{align}
\vec{z}_{u} &= \vec{W}_{\text{soft}}\vec{c}_{u} + \vec{b}_{\text{soft}}, \nonumber \\
\vec{y}_{u} &= \text{Softmax}(\vec{z}_{u}). \label{eq:CTCAttn-generate}
\end{align}
To include non-uniform attention weights $\alpha_{u,t}$ instead of uniform weights ($\alpha_{u,t} = \frac{1}{C} $ in Eq. \eqref{eq:CTCAttn-TimeConvolution}), we use the Attend(.) function,
\begin{align}
\bm\alpha_{u} &= \text{Attend}(\vec{z}_{u-1}, \bm{\alpha}_{u-1}, \vec{g}). \label{eq:CTCAttn-attend}
\end{align}
Thus, Eq. \eqref{eq:CTCAttn-attend} represents hybrid attention (HA) as it encodes both content ($\vec{z}_{u-1}$) and location ($\bm{\alpha}_{u-1}$) information. In the absence of $\bm{\alpha}_{u-1}$, Eq. \eqref{eq:CTCAttn-attend} would represent content attention (CA).

The performance of the attention model can be improved further by providing more reliable content information. This is possible by introducing another recurrent network that can utilize content from several time steps in the past. This network, in essence, would learn an implicit language model (LM) and can be represented as,
\vspace{-2mm}
\begin{align}
\vec{z}^{\text{LM}}_{u-1} &= \mathcal{H}(\vec{x}_{u-1}, \vec{z}^{\text{LM}}_{u-2}), \quad
\vec{x}_{u-1} = 
\begin{bmatrix}
\vec{z}_{u-1} \\
\vec{c}_{u-1}
\end{bmatrix}, \label{eq:CTCAttnLM-LSTM} \\
\bm\alpha_{u} &= \text{Attend}(\vec{z}^{\text{LM}}_{u-1}, \bm{\alpha}_{u-1}, \vec{g}),  \label{eq:CTCAttn-attendLM}
\end{align}
where $\mathcal{H}(.)$ is a LSTM unit.

In the final step to improve attention, each of the $n$ components of $\vec{g}_{t} \in \setsym{R}^{n}$ in Eq. \eqref{eq:CTCAttn-TimeConvolution} could be weighted distinctively. This is possible by replacing the scalar attention weight $\alpha_{u,t} \in [0, 1]$ with a vector attention weight $\bm\alpha_{u,t} \in [0, 1]^{n}$ for each $t \in [u-\tau,\ u+\tau]$. 
Under this formulation, the context vector $\vec{c}_{u}$ can be computed using,
\vspace{-2mm}
\begin{align}
\vec{c}_{u} &= \gamma \sum_{t=u-\tau}^{u+\tau} \bm\alpha_{u,t} \odot \vec{g}_{t}, \label{eq:CTCAttn-Comp-annotate}
\end{align}
where $\odot$ is the Hadamard product. 

\vspace{-2mm}
\subsection{Comparison with Other End-to-end Methods}\vspace{-2mm}
\label{ssec: CompareE2E}

In addition to CTC, there are also popular E2E methods in ASR, such as RNN encoder-decoder (RNN-ED) \cite{Chan-LAS, lu2016training}  and  RNN transducer (RNN-T) \cite{rao2017exploring}. Initially working on letter units, these methods recently got significant improvement when working on word-piece units \cite{schuster2012japanese}, either pre-trained \cite{chiu2017state, rao2017exploring} or automatically derived \cite{chan2016latent} during training. In all these works, all the words are decomposed into word-piece units which range from single letter all the way up to entire words. In contrast, our acoustic-to-word model directly uses frequent words as basic units, and only decomposes infrequent words into a sequence of frequent words and multi-letters. The majority units are still words. Therefore, our units are more stable and natural for the E2E system outputting word hypotheses. In \cite{lu2016training}, words were also used as the basic units with the RNN-ED structure. However, the reported WER was much higher than the one obtained with traditional systems. 

As extensions of CTC, both RNN-T and RNN aligner \cite{sak2017recurrent} either change the objective function or the training process to relax the frame independence assumption of CTC. The proposed attention CTC in Section \ref{ssec: CTCAttn} is another solution by working on hidden layer representation with more context information without changing the CTC objective function and training process. 

\vspace{-2mm}
\section{Experiments}\vspace{-2mm}
\label{sec: Expts}

The proposed methods were evaluated using the Microsoft's Cortana voice assistant task.  The training dataset contains approximately 3.3 million short utterances ($\sim$ 3400 hours) in US-English. The test set contains about 5600 utterances ($\sim$ 6 hours). The base feature vector for every 10 ms is a 80-dimensional vector containing log filterbank energies. The base feature vectors in three continuous frames are stacked together as the 240-dimension input feature to the CTC models \cite{sak2015fast}. All CTC models are bi-directional LSTM models.

We first built a phoneme-based bi-directional 6-layer LSTM model trained with the CTC criterion, modeling around 9000 tied context-dependent (CD) phonemes.  Every layer of the bi-directional LSTM has 512 memory units in each direction. Unless otherwise stated, all CTC models except attention CTC models in this study use the same structure as this model. This CD-phone CTC model has 9.28\% WER 
when decoding with a 5-gram LM with totally around 100 million (M) n-grams. In this study, except this CD-phone CTC model, all the other CTC models are E2E models using greedy decoding which generate the final output sequence without using any LM or complicated decoding process.

Next, we built an acoustic-to-word CTC model with the same model structure as the CD-phone CTC by modeling around 27k most frequent words in the training data. These frequent words occurred at least 10 times in the training data. All other infrequent words were mapped to an OOV output token.  We have also tried other word-based CTCs with varying number of output units. However, the model using 27k word outputs performs the best. This word-based LSTM-CTC model yields 9.84\% WER, among which the OOV tokens contribute 1.87\% WER. It significantly improves the WER of uni-directional word-based CTC reported in \cite{Li17CTCnoOOV} which indicates the bi-directional modeling is critical to the E2E system.  


\vspace{-2mm}
\subsection{Letter CTC with Attention}\vspace{-2mm}
\label{ssec: exp_letter}
As the word output in the letter-based CTC is used to replace the OOV token from the word-based CTC model during testing,  the letter-based CTC should be as accurate as possible. In this set of experiments, we first evaluate the impact of using different size of letter units for the vanilla CTC \cite{Graves-CTCFirst}. All the letter-based CTC models are 6-layer bi-directional LSTM models. The single-letter set has 30 symbols, including 26 English characters [a-z], ', *, \$, and blank. The double-letter and triple-letter sets have 763 and 8939 symbols respectively, covering all the double-letter and triple-letter occurrence in the training set. As shown in the second column of Table \ref{Tab:WER_letterCTC}, the WER reduces significantly when the output units become larger, i.e., more stable. The letter-based CTC using triple-letter as output units achieves 13.28\% WER, reducing 24.29\% relative WER from the letter-based CTC using single-letter as output units. 

The attention CTC presented in Section \ref{ssec: CTCAttn} is then trained with $\tau$ empirically set as 4 (context window size $C$ = 9). As shown in the third column of  Table \ref{Tab:WER_letterCTC}, attention CTC improves the vanilla CTC hugely, obtaining 18.47\%, 20.88\%, and 14.46\% relative WER reduction for single-letter, double-letter, and triple-letter CTC models, respectively. The best letter-based E2E CTC model is the one with triple-letter outputs and attention modeling, which can obtain 11.36\% WER. 

The hybrid CTC model described in Section \ref{ssec: hybCTC} has both word-based CTC and letter-based CTC, which share 5 hidden LSTM layers.
On top of the shared hidden layers, we add a new LSTM hidden layer and a softmax layer to model letter (single, double, or triple-letters) outputs. Attention modeling is applied to boost the performance. As shown in the fourth column of Table   \ref{Tab:WER_letterCTC}, the WER of letter-based CTC with such shared-hidden-layer constraint performs worse than its counterpart. This indicates one shortcoming of the hybrid CTC -- it sacrifices the accuracy of the letter-based CTC	because of the shared-hidden-layer constraint used to synchronize the word outputs between the word-based and letter-based CTC. 


\begin{table}[!t]
\centering
\caption{WERs of letter-based CTC models with single, double, and triple-letter output units. Three structures are evaluated: vanilla CTC \cite{Graves-CTCFirst}, attention CTC, and attention CTC sharing 5 hidden layers with the word CTC.}
\vspace{-3mm}
\begin{tabular}{l|c c c}
   \hline
E2E Model  & \multicolumn{3}{c}{WER (\%)} \\
         		   						  &Vanilla      &Attention    &Attention \\ 
														&      &    &5 layers sharing \\\hline
single-letter						    &17.54 &14.30    & 16.74   \\
double-letter            &15.37 &12.16   &14.00  \\
triple-letter            &13.28 &11.36   &12.81  \\ \hline
\end{tabular}
\label{Tab:WER_letterCTC}
\end{table}

\vspace{-2mm}
\subsection{Hybrid CTC}\vspace{-2mm}
\label{ssec: exp_hybrid}

As the CTC models with double-letter and triple-letter output units worked very well in Table \ref{Tab:WER_letterCTC}, we use them to build the hybrid CTC models with the OOV lookup process described in Section \ref{ssec: hybCTC}. Both hybrid models achieved 9.66\% WER as shown in Table \ref{Tab:WER_HybCTC}. Several factors contribute to such small improvement (from 9.84\% WER of the word-based CTC) of the hybrid CTC. First, the shared-hidden-layer constraint degrades the performance of the letter-based CTC, potentially affecting the final hybrid system performance. Second, although the shared-hidden-layer constraint helps to synchronize the word outputs from the word and letter based CTC, we still observed that the time synchronization can fail sometimes. In such cases, the OOV token is replaced with its neighboring frequently occurring word because of word segments misalignment. Because of these factors, although the triple-letter CTC is better than double-letter CTC in Table \ref{Tab:WER_letterCTC}, there is no difference when they are combined with the baseline word CTC in the hybrid CTC setup in which they only handle the small portion of OOV words.
 

\begin{table}[!t]
\centering
\caption{WERs of vanilla word-based CTC and hybrid CTC models. All Hybrid CTC models have a word-based CTC and a letter-based attention CTC, sharing 5 hidden layers.}
\vspace{-3mm}
\begin{tabular}{l|c}
   \hline
E2E Model  & \multicolumn{1}{c}{WER (\%)} \\
 \hline
Word-based CTC 						    &9.84       \\ \hline
Word-based CTC + double-letter Attention CTC            &9.66     \\
Word-based CTC + triple-letter Attention CTC              &9.66     \\
 \hline
\end{tabular}
\vspace{-6mm}
\label{Tab:WER_HybCTC}
\end{table}

\vspace{-2mm}
\subsection{CTC with Mixed Units}\vspace{-2mm}
\label{ssec: exp_mix}

We evaluate the CTC with mixed units in Table \ref{Tab:WER_mixCTC}. In the first experiment, the mixed units contain single-letters and 27k frequent words. During training, OOV words are decomposed into single-letter sequence.  As analyzed in Section \ref{ssec: mixCTC}, artificially decomposing OOV words into letter sequence while keeping the frequent words confuses CTC training for these types of words. Therefore, the trained CTC model achieved 20.10\% WER. When looking at the posterior spikes of this model, we observed that the word spikes and letter spikes are scattered into each other which proves our hypothesis. 

Next, we decompose OOV words into frequent word and single-letter sequences, and train the CTC network with the mixed units (around 27k). Immediately, the WER improved to 10.17\%, but still a little worse than the baseline word-based CTC. This is because the single-letter sequence brings instability to the modeling. 
When we decompose the OOV words into frequent words and double-letters (totally 27k units), the situation becomes better, and the resulting WER is 9.58\%. When the triple-letters and frequent words are used (totally 33k units), the WER reaches 9.32\%, beating the baseline word-based CTC by  5.28\% relative WER reduction. 

Finally, we  improve the final E2E CTC model by applying attention CTC. To save computational cost with large number of output units, we didn't integrate the implicit LM in Eq.\eqref{eq:CTCAttnLM-LSTM}. The WER becomes 8.65\%, which is about relative 12.09\% WER reduction from the 9.84\% WER of vanilla word-based CTC. Such a model without using LM and complex decoder also outperforms the traditional context-dependent-phoneme CTC with strong LM and decoder which obtained 9.28\% WER. Note that the proposed method not only reduces the WER of the word-based CTC, but also improves the user experience. The proposed method provides more meaningful output without outputting  any OOV token to distract users. Most of the time, even if the proposed method cannot get the OOV word right, it comes out with a very close output. For example,  the proposed method recognize ``text fabine'' as ``text fabian'' and ``call zubiate'' as ``call zubiat'', while the vanilla word-based CTC can only output ``text OOV'' and ``call OOV''.

\begin{table}[!t]
\centering
\caption{WERs of the vanilla word-based CTC and the CTC with mixed units.}
\vspace{-3mm}
\begin{tabular}{l|c}
   \hline
E2E Model  & \multicolumn{1}{c}{WER (\%)} \\
 \hline
word-based CTC 						    &9.84       \\ \hline
mixed (OOV: single-letter) CTC &20.10 \\ 
mixed(OOV: word + single-letter) CTC            &10.17     \\
mixed (OOV: word + double-letter)  CTC            &9.58    \\
mixed (OOV: word + triple-letter)  CTC            &9.32     \\
mixed (OOV:  word + triple-letter) attention CTC              &8.65     \\
 \hline
\end{tabular}
\vspace{-4mm}
\label{Tab:WER_mixCTC}
\end{table}

\vspace{-3mm}
\section{Conclusions}\vspace{-3mm}
\label{sec: Conclusions}
We advance acoustic-to-word CTC model with a mixed-unit CTC whose output units are frequent words combined with sequences of multi-letters. For the frequent word, we just model it with a unique output node. For the OOV word, we decompose it into a sequence of frequent words and multi-letters. We present the attention CTC which significantly improves the modeling power of CTC. The proposed method is simpler and more effective than the hybrid CTC which has to rely on shared-hidden-layer to maintain the time synchronization of word outputs between the word-based and letter-based CTCs. We evaluate all these methods on a 3400 hours Microsoft Cortana voice assistant task. The proposed acoustic-to-word CTC with mixed-units reduces relative 5.28\% WER from the vanilla word-based CTC, and reduces relative 12.09\% WER if combined with the proposed attention CTC. Such an acoustic-to-word CTC is a pure end-to-end model without any LM and complex decoder. It also outperforms the traditional context-dependent-phoneme CTC with strong LM and decoder by relative 6.79\% WER reduction.

%
\clearpage
\bibliographystyle{IEEEbib}
\balance
\bibliography{strings,refs}

\begin{thebibliography}{10}

\bibitem{Graves-CTCFirst}
A.~Graves, S.~Fern\'{a}ndez, F.~Gomez, and J.~Schmidhuber,
\newblock ``{Connectionist Temporal Classification: Labelling Unsegmented
  Sequence Data with Recurrent Neural Networks},''
\newblock in {\em Proc. Int. Conf. in Learning Representations}, 2006, pp.
  369--376.

\bibitem{Graves-E2EASR}
A.~Graves and N.~Jaitley,
\newblock ``{Towards End-to-End Speech Recognition with Recurrent Neural
  Networks},''
\newblock in {\em Proc. of Machine Learning Research}, 2014, pp. 1764--1772.

\bibitem{Hannun-DeepSpeech}
A.~Y. Hannun, C.~Case, J.~Casper, B.~Catanzaro, G.~Diamos, E.~Elsen,
  R.~Prenger, S.~Satheesh, S.~Sengupta, A.~Coates, and A.~Y. Ng,
\newblock ``{Deep Speech: Scaling up End-to-End Speech Recognition},''
\newblock {\em CoRR}, vol. abs/1412.5567, 2014.

\bibitem{sak2015learning}
H.~Sak, A.~Senior, K.~Rao, O.~Irsoy, A.~Graves, F.~Beaufays, and J.~Schalkwyk,
\newblock ``{Learning Acoustic Frame Labeling for Speech Recognition with
  Recurrent Neural Networks},''
\newblock in {\em Proc. ICASSP}, 2015, pp. 4280--4284.

\bibitem{sak2015fast}
H.~Sak, A.~Senior, K.~Rao, and F.~Beaufays,
\newblock ``{Fast and Accurate Recurrent Neural Network Acoustic Models for
  Speech Recognition},''
\newblock in {\em Proc. Interspeech}, 2015.

\bibitem{miao2015eesen}
Y.~Miao, M.~Gowayyed, and F.~Metze,
\newblock ``{EESEN: End-to-End Speech Recognition using Deep {RNN} Models and
  {WFST}-based Decoding},''
\newblock in {\em Proc. ASRU}, 2015, pp. 167--174.

\bibitem{kanda2016maximum}
Naoyuki Kanda, Xugang Lu, and Hisashi Kawai,
\newblock ``{Maximum a posteriori Based Decoding for CTC Acoustic Models.},''
\newblock in {\em Proc. Interspeech}, 2016, pp. 1868--1872.

\bibitem{soltau2016neural}
H.~Soltau, H.~Liao, and H.~Sak,
\newblock ``{Neural Speech Recognizer: Acoustic-to-word LSTM Model for Large
  Vocabulary Speech Recognition},''
\newblock {\em arXiv preprint arXiv:1610.09975}, 2016.

\bibitem{Zweig-AdvancesNeuralASR}
G.~Zweig, C.~Yu, J.~Droppo, and A.~Stolcke,
\newblock ``{Advances in All-Neural Speech Recognition},''
\newblock in {\em Proc. ICASSP}, 2017, pp. 4805--4809.

\bibitem{liu2017gram}
H.~Liu, Z.~Zhu, X.~Li, and S.~Satheesh,
\newblock ``{Gram-CTC: Automatic Unit Selection and Target Decomposition for
  Sequence Labelling},''
\newblock {\em arXiv preprint arXiv:1703.00096}, 2017.

\bibitem{audhkhasi2017direct}
K.~Audhkhasi, B.~Ramabhadran, G.~Saon, M.~Picheny, and D.~Nahamoo,
\newblock ``{Direct Acoustics-to-Word Models for English Conversational Speech
  Recognition},''
\newblock {\em arXiv preprint arXiv:1703.07754}, 2017.

\bibitem{Li17CTCnoOOV}
J.~Li, G.~Ye, R.~Zhao, J.~Droppo, and Y.~Gong,
\newblock ``{Acoustic-to-Word Model Without OOV},''
\newblock in {\em Proc. ASRU}, 2017.

\bibitem{Yu-RecentProgDeepLearningAcousticModels}
D.~Yu and J.~Li,
\newblock ``{Recent Progresses in Deep Learning Based Acoustic Models},''
\newblock {\em IEEE/CAA J. of Autom. Sinica.}, vol. 4, no. 3, pp. 399--412,
  July 2017.

\bibitem{Li2018Speaker}
J.~Li, R.~Zhao, et~al.,
\newblock ``Developing far-field speaker system via teacher-student learning,''
\newblock in {\em Proc. ICASSP}, 2018.

\bibitem{bazzi2002modelling}
Issam Bazzi,
\newblock {\em Modelling out-of-vocabulary words for robust speech
  recognition},
\newblock Ph.D. thesis, Massachusetts Institute of Technology, 2002.

\bibitem{decadt2002transcription}
Bart Decadt, Jacques Duchateau, Walter Daelemans, and Patrick Wambacq,
\newblock ``Transcription of out-of-vocabulary words in large vocabulary speech
  recognition based on phoneme-to-grapheme conversion,''
\newblock in {\em Proc. ICASSP}, 2002, vol.~1, pp. I--861.

\bibitem{yazgan2004hybrid}
Ali Yazgan and Murat Saraclar,
\newblock ``Hybrid language models for out of vocabulary word detection in
  large vocabulary conversational speech recognition,''
\newblock in {\em Proc. ICASSP}, 2004, vol.~1, pp. I--745.

\bibitem{bisani2005open}
Maximilian Bisani and Hermann Ney,
\newblock ``Open vocabulary speech recognition with flat hybrid models.,''
\newblock in {\em Proc. Interspeech}, 2005, pp. 725--728.

\bibitem{audhkhasi2017building}
Kartik Audhkhasi, Brian Kingsbury, Bhuvana Ramabhadran, George Saon, and
  Michael Picheny,
\newblock ``Building competitive direct acoustics-to-word models for english
  conversational speech recognition,''
\newblock in {\em submitted to ICASSP}, 2018.

\bibitem{Hochreiter1997long}
S.~Hochreiter and J.~Schmidhuber,
\newblock ``{Long Short-Term Memory},''
\newblock {\em Neural computation}, vol. 9, no. 8, pp. 1735--1780, 1997.

\bibitem{Graves2013speech}
A.~Graves, A.~Mohamed, and G.~Hinton,
\newblock ``Speech recognition with deep recurrent neural networks,''
\newblock in {\em Proc. ICASSP}, 2013, pp. 6645--6649.

\bibitem{Sak2014long}
H.~Sak, A.~Senior, and F.~Beaufays,
\newblock ``Long short-term memory recurrent neural network architectures for
  large scale acoustic modeling.,''
\newblock in {\em Proc. Interspeech}, 2014, pp. 338--342.

\bibitem{siohan2017ctc}
Olivier Siohan,
\newblock ``{CTC} training of multi-phone acoustic models for speech
  recognition,''
\newblock in {\em Proc. Interspeech}, 2017, pp. 709--713.

\bibitem{Das18CTCAttention}
A.~Das, J.~Li, R.~Zhao, and Y.~Gong,
\newblock ``Advancing connectionist temporal classification with attention
  modeling,''
\newblock in {\em Proc. ICASSP}, 2018.

\bibitem{Chan-LAS}
W.~Chan, N.~Jaitly, Q.~V. Le, and O.~Vinyals,
\newblock ``{Listen, Attend and Spell},''
\newblock {\em CoRR}, vol. abs/1508.01211, 2015.

\bibitem{lu2016training}
Liang Lu, Xingxing Zhang, and Steve Renais,
\newblock ``On training the recurrent neural network encoder-decoder for large
  vocabulary end-to-end speech recognition,''
\newblock in {\em Proc. ICASSP}, 2016, pp. 5060--5064.

\bibitem{rao2017exploring}
Kanishka Rao, Ha{\c{s}}im Sak, and Rohit Prabhavalkar,
\newblock ``Exploring architectures, data and units for streaming end-to-end
  speech recognition with {RNN}-transducer,''
\newblock in {\em Proc. ASRU}, 2017.

\bibitem{schuster2012japanese}
Mike Schuster and Kaisuke Nakajima,
\newblock ``Japanese and korean voice search,''
\newblock in {\em Proc. ICASSP}, 2012, pp. 5149--5152.

\bibitem{chiu2017state}
Chung-Cheng Chiu, Tara~N Sainath, et~al.,
\newblock ``{State-of-the-art speech recognition with sequence-to-sequence
  models},''
\newblock in {\em submitted to ICASSP}, 2018.

\bibitem{chan2016latent}
William Chan, Yu~Zhang, Quoc Le, and Navdeep Jaitly,
\newblock ``Latent sequence decompositions,''
\newblock {\em arXiv preprint arXiv:1610.03035}, 2016.

\bibitem{sak2017recurrent}
Hasim Sak, Matt Shannon, Kanishka Rao, and Fran{\c{c}}oise Beaufays,
\newblock ``{Recurrent neural aligner: An encoder-decoder neural network model
  for sequence to sequence mapping},''
\newblock in {\em Proc. Interspeech}, 2017.

\end{thebibliography}

\end{document}